\documentclass[letterpaper, 10 pt, journal, twoside]{IEEEtran}
\IEEEoverridecommandlockouts                              

\usepackage[utf8]{inputenc} 
\usepackage{amsmath}
\usepackage{amssymb}
\usepackage{booktabs}
\usepackage{graphicx}
\usepackage{balance}
\usepackage{biblatex}

\title{GesGPT: Speech Gesture Synthesis With Text Parsing from ChatGPT}

\author{Nan Gao$^{1}$, Zeyu Zhao$^{2}$, Zhi Zeng$^{3}$, Shuwu Zhang$^{3}$, Dongdong Weng$^{4}$ and Yihua Bao$^{4}$
\thanks{Manuscript received: September 18, 2023; Revised: November 16, 2023; Accepted: January 15, 2024.}
\thanks {This paper was recommended for publication by
Editor Gentiane Venture upon evaluation of the Associate Editor and Reviewers’ comments.

This work was supported by the National Key R\&D Program of China under Grant 2022YFF0902202.}
\thanks{$^{1}$First Author is with the Institute of Automation, Chinese Academy of Sciences, Beijing, China.
        {\tt\small nan.gao@ia.ac.cn}}%
\thanks{$^{2}$Second Author is with University of Chinese Academy of Sciences, Beijing, China, and also with the Institute of Automation, Chinese Academy of Sciences, Beijing, China.
        {\tt\small zhaozeyu2019@ia.ac.cn}}%
\thanks{$^{3}$Third Author and Fourth Author are with Beijing University of Posts and Telecommunications, Beijing, China.             
        {\tt\small zhi.zeng@bupt.edu.cn; shuwu.zhang@bupt.edu.cn}}%
\thanks{$^{4}$Fifth Author and Sixth Author are with Beijing Engineering Research Center of Mixed Reality and Advanced Display, Beijing, China, and also with the Institute of Technology, Beijing, China.
        {\tt\small crgj@bit.edu.cn; boye1900@outlook.com}%
}%
\thanks{Digital Object Identifier (DOI): see top of this page.}
}

\begin{document}

\maketitle
\begin{abstract}

Gesture synthesis has gained significant attention as a critical research field, aiming to produce contextually appropriate and natural gestures corresponding to speech or textual input. Although deep learning-based approaches have achieved remarkable progress, they often overlook the rich semantic information present in the text, leading to less expressive and meaningful gestures. In this letter, we propose GesGPT, a novel approach to gesture generation that leverages the semantic analysis capabilities of large language models , such as ChatGPT. By capitalizing on the strengths of LLMs for text analysis, we adopt a controlled approach to generate and integrate professional gestures and base gestures through a text parsing script, resulting in diverse and meaningful gestures. Firstly, our approach involves the development of prompt principles that transform gesture generation into an intention classification problem using ChatGPT. We also conduct further analysis on emphasis words and semantic words to aid in gesture generation. Subsequently, we construct a specialized gesture lexicon with multiple semantic annotations, decoupling the synthesis of gestures into professional gestures and base gestures. Finally, we merge the professional gestures with base gestures. Experimental results demonstrate that GesGPT effectively generates contextually appropriate and expressive gestures.

\end{abstract}

\section{INTRODUCTION}

Gesture synthesis is a research field that aims to create natural and contextually appropriate gestures that correspond to given speech or textual input. The majority of existing methods adopt deep learning-based approaches, which primarily focus on audio features extracted from speech signals to model and generate gestures \cite{1}\cite{2}. Several of these approaches treat gesture synthesis as a regression problem and utilize various architectures, such as convolutional neural networks \cite{3}\cite{4}, sequence networks \cite{5}\cite{6}, and generative models \cite{7}\cite{8}, to learn relationships between speech and gestures.

However, these existing methods have certain limitations. Firstly, they often overlook the rich semantic information embedded in textual input, which could contribute to the generation of more expressive and meaningful gestures. For instance, an analysis of the GENEA Challenge 2022 \cite{9} has shown that the generated gestures exhibit superior human-like similarity compared to motion capture data. However, enhancing the semantic expressiveness of these gestures still requires further exploration. Secondly, the deep learning-based methods for gesture synthesis tend to yield average results, often failing to generate nuanced and intricate hand movements \cite{10}. Lastly, they may not adequately address the inherent intention in gesture expressions, potentially leading to the generation of less plausible or contextually inappropriate gestures. Given these limitations, it is crucial to explore alternative research approaches that can leverage the wealth of information available in textual input and effectively model gesture-speech associations.

Large Language Models (LLMs), such as GPT-3 \cite{11} and BERT \cite{12}, have demonstrated remarkable capabilities in semantic analysis, contextual understanding, and extracting meaningful information from text. These models have been applied to various natural language processing tasks with impressive results \cite{13}. Recently, several studies have successfully employed ChatGPT in the field of robotics \cite{14}\cite{15}\cite{16}. Given the potential applications of gestures in both agent and robotic domains \cite{17}\cite{18}, it is essential to explore the utilization of LLMs in the gesture generation field.

By leveraging the robust semantic analysis capabilities of LLMs, we introduce GesGPT, a novel approach to gesture generation that focuses on text parsing using ChatGPT. While previous studies using deep learning techniques have successfully generated realistic gestures from text or speech input, these gestures often lack semantic coherence. To address this limitation, our approach incorporates prompt engineering design principles with ChatGPT to generate expressively coherent gestures from text input. Additionally, deep learning methods have proven effective in modeling the rhythmic attributes between speech and gestures \cite{19}. Therefore, we explore the integration of existing deep learning frameworks with LLMs to mutually enhance gesture generation. Following a `human-on-the-loop' paradigm, we generate semantic scripts through prompt engineering and combine them with gestures derived from deep learning methodologies.

In summary, our contributions are as follows:
\begin{itemize}

\item We introduce GesGPT, a gesture generation framework that employs ChatGPT for text parsing to generate gestures with semantic meaning. Drawing inspiration from gesture cognition research, we break down text parsing into several tasks, such as intent classification, emphasis word identification, and semantic word recognition. This approach enables us to derive parsing scripts that guide gesture generation, resulting in a controllable and editable approach for generating gestures.

\item Co-speech gesture can be considered as composed of a series of basic gesture units \cite{20}. Inspired by this, we decompose gesture generation into two components: professional gestures and base gestures related to rhythm. Professional gestures refer to complete sequences of gesture units. Based on this, we have constructed a comprehensive specialized gesture lexicon with semantic annotations derived from video data. By integrating the gesture lexicon into the gesture generation framework, along with text parsing and utilizing a script-based search module, our method generates expressive gestures.

\item We conducted extensive experiments on both English and Chinese datasets. The experimental results reveal that GesGPT effectively capitalizes on the strengths of LLMs in understanding the inherent meaning and context of text input. As a result, it produces contextually appropriate and expressive gestures that enrich the overall communication experience.

\end{itemize}

\begin{figure*}[htbp]
\begin{center}
\includegraphics[width=0.7\linewidth]{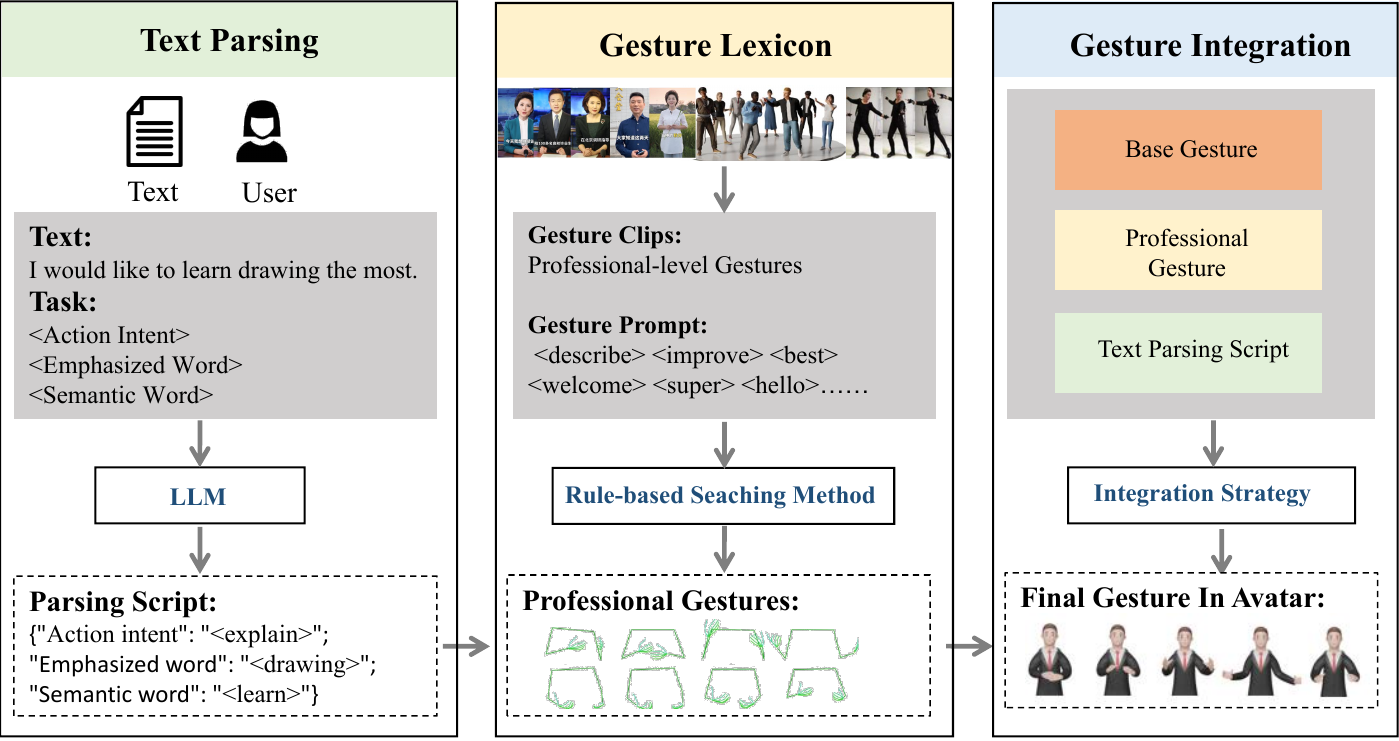}
\end{center}
\caption{The Pipeline of GesGPT. Firstly, the Text Parsing module is employed to generate a gesture script from the input text. Then, the Gesture Lexicon module is used to search and retrieve corresponding professional gestures based on the script. Next, the Gesture Integration module combines the semantically professional gestures from the gesture lexicon with rhythmic base gestures. }
\label{fig:1}
\end{figure*}

\section{Related Work}

\subsection{Gesture Synthesis}
Recent research has shown that modeling gestures using a diffusion model \cite{21}\cite{8} or VQ-VAE \cite{22}\cite{7} can produce natural and diverse gestures. In the field of cognitive psychology \cite{23}, researchers have discovered that speech and gesture can enhance each other's comprehension when they convey information that is semantically consistent. For modeling semantic gestures, manually designed rules have been shown to effectively preserve semantics within limited domains \cite{24}\cite{25}. Various deep learning approaches have been combined with rule-based methods \cite{26}\cite{27} or integrated with predefined gesture dictionaries \cite{28}, to produce more natural and semantically rich gestures. Additionally, Seeg \cite{29} explored the use of a gesture semantic classifier to obtain semantic labels for gestures and ensure coherency in generated semantics. Teshima et al. \cite{30} extensively categorized gesture types into beat, imagistic, and no-gesture following McNeill's work \cite{31} and modeled them separately. In our approach, we refer to McNeill's \cite{31} classification of gesture functions to classify text into multiple intents, and utilize text parsing scripts derived from ChatGPT to generate expressive gestures with communicative functionality.

\subsection{LLMs in Embodied Artificial Intelligence}

ChatGPT for Robotics \cite{15} discusses the application of the ChatGPT model, which is based on natural language processing in robotics. By incorporating prompt engineering design principles and creating a sophisticated function library, ChatGPT can be adapted to various robotic tasks and simulators, demonstrating its potential in the robotics domain. However, while LLMs possess strong analytical capabilities for existing knowledge, they face challenges in explaining non-linguistic environments such as physical settings. To address this, Huang et al.'s work \cite{32} utilize semantic knowledge from language models and consider external environmental factors by constructing action sequences. This foundational model integrates language models and environmental factors to achieve action-based reasoning. VoxPoser \cite{33} proposes a framework for mapping language instructions to trajectories for robot operations, achieving impressive results across multiple robot manipulation tasks.These works highlight the immense potential of introducing LLMs like ChatGPT into the field of embodied intelligence. Additionally, MotionGPT \cite{34} proposes a unified motion-language model, which is pre-trained based on an motion vocabulary . This approach demonstrates promising results in various tasks, including motion prediction. This paper aims to explore a method for synthesizing co-speech gestures using text parsing results based on LLMs.

\section{GesGPT}

\subsection{Gesture Synthesis Formulation}


Deep learning-based methods typically perform gesture synthesis by dividing the process into multiple segments. Given the segmentation of the original speaker's videos, we obtain $N$ video segments, each comprising $K$ frames. Body landmarks are extracted to form gesture segments $G_{i=1}^{N} = \{G_{1}, ..., G_{N}\}$, where $G_{i}\in \mathbb{R}^{K*D_{G}}$, and $D_{G}$ represents the corresponding dimension. Similarly, the audio and text are temporally segmented to form audio segments $S_{i=1}^{N}=\{S_{1}, ..., S_{N}\}$ and text segments $T_{i=1}^{N}=\{T_{1}, ..., T_{N}\}$, where $S_{i}\in \mathbb{R}^{K*D_{S}}$ and $T_{i}\in \mathbb{R}^{K*D_{T}}$. Consequently, the deep learning-based gesture generation method, denoted as $M$, can be expressed as $G_{i} = M(S_{i}, T_{i})$.

Current deep learning methods for gesture generation predominantly treat gesture synthesis as a regression problem, primarily focusing on audio input. Nevertheless, text encompasses rich semantic information. In this paper, we propose to approach gesture generation as a text classification and recognition problem. We further explore the application of LLMs for text parsing. We generate gestures grounded on the parsing script derived from purposefully designed prompts.

\subsection{Our Method}

\subsubsection{Overall Framework}

As shown in Fig.1 , our approach consists of three modules: the text parsing module that leverages LLMs such as ChatGPT, the professional gesture lexicon module with semantic annotations, and the gesture integration module based on the text parsing results. Initially, we analyze the text by applying parsing principles derived from the literature on gesture cognition and our observations of speech-style videos. This step is facilitated by utilizing pre-designed prompts in ChatGPT, which enables us to acquire text parsing results. Following the parsing of the text, the pre-defined search rules are employed to match the corresponding professional gestures from the gesture lexicon. Subsequently, within the gesture integration module, diverse gestures are integrated based on the parsing script. This framework allows for the controlled generation of meaningful and visually comprehensive gestures in accordance with the text parsing script.

\subsubsection{Text Parsing from ChatGPT}

Speech gestures can enhance speakers' ability to effectively convey information and emphasize essential points during presentations. Our research primarily focuses on exploring the utilization of LLMs to assist in generating gestures with enriched semantic expression. This process is depicted in Fig.2. Firstly, we process one sentence at a time and divide it into three subtasks: (i) determining the intention expressed in the sentence, (ii) identifying the words that require emphasis within the sentence, and (iii) recognizing the words that carry semantic significance. We design specific prompts for these three objectives, with expressing intention defined as a classification task, emphasizing words and semantic words defined as recognition tasks. Leveraging ChatGPT, we can analyze the deep semantic information embedded within the sentence. The obtained text parsing script using this approach enables the controlled generation of co-speech gestures that exhibit significant expressiveness.

\begin{figure*}[htbp]
\begin{center}
\includegraphics[width=0.7\linewidth]{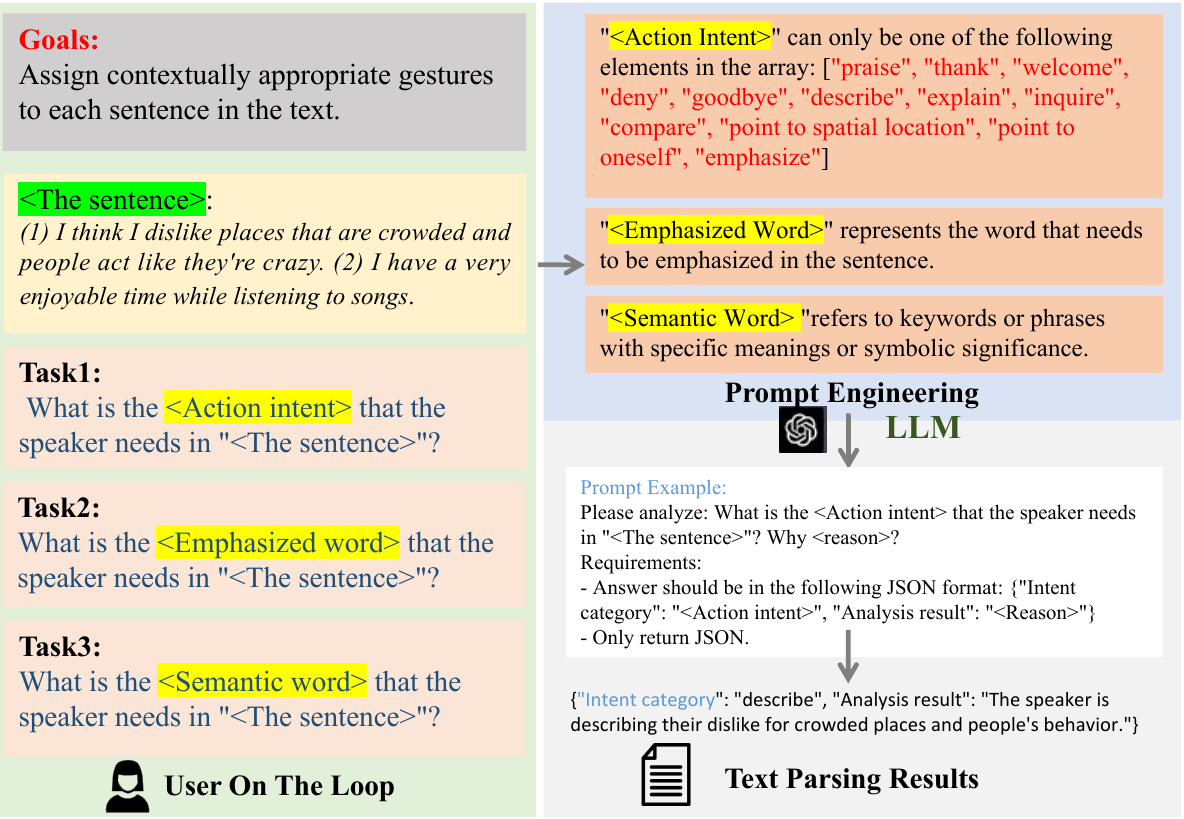}
\caption{The Pipeline of Text Parsing. It involves three constituent subtasks: action intent classification, emphasis word recognition, and semantic word recognition. Finally, the results are integrated into a parsing script.}
\end{center}
\label{fig:1}
\end{figure*}

\textbf{Action Intention:}
To generate expressive and meaningful gestures, we classify them based on their intended meaning using McNeill's work \cite{31} as a framework. McNeill categorizes gestures into five types: emblematic, iconic, metaphoric, deictic, and beat gestures. Building on this, we further divide these gesture types into several categories based on different speaking intents, including praise, thank, welcome, deny, goodbye, describe, explain, inquire, compare, point to spatial location, point to oneself, and emphasize. For instance, the iconic gesture is used to convey what is being said, and we define the corresponding textual intention as `describe'. Metaphoric gestures that illustrate abstract concepts are categorized as `explain,' `inquire,' or `compare' in terms of textual intention. Subsequently, we treat the determination of textual intent as a classification task. When given a sentence as input, we prompt the LLMs to infer the most appropriate intention category, denoted as  $label\_1$  from this set of classification labels. The LLMs then provide a response in the format of and provide a response in the form of $<intention:label\_1>$.

\textbf{Emphasized Word:}
The timing of co-speech gestures varies among individuals. In this article, we make the assumption that each sentence corresponds to a specific gesture. We prompt the LLMs to identify the keyword, denoted as $<stroke:label\_2>$, from the sentence. This allows us to insert gestures from the gesture lexicon based on the position of emphasized words, ensuring the coherence and appropriateness of generated gestures.

\textbf{Semantic Word:}
We have observed a correlation between certain words and gestures. For example, when the word `excellent' is mentioned, a thumbs-up gesture is often made. We refer to these words as semantic words and prompt the LLMs to identify the semantic word, denoted as $<semantic:label\_3>$. Semantic gestures are characterized by specific meanings or symbolic significance. Parsing out semantic words from the text will aid in generating gestures that serve as effective auxiliary expressions.

In the text parsing module, for a sentence $T$, we obtain the parsing results of the three corresponding tasks, namely $\{T: <intention:label\_1>,<stroke:label\_2>,<semantic:label\_3>\}$. Subsequently, we organize all the parsed results of the text into a JSON-based parsing script, which serves as the foundation for gesture generation and integration based on the script.

\subsubsection{Gesture Lexicon}According to Kendon's research \cite{35}, a gesture can be divided into five stages in the temporal dimension, namely rest position, preparation, stroke, hold, and retraction. Based on this, we define the basic unit in the gesture lexicon as a gesture with the aforementioned initial and final stages. Specifically, a gesture in the lexicon starts from a rest position, goes through a series of hand movements, and ends at the rest position, as illustrated in Fig.3.
\begin{figure}[htbp]
\begin{center}
\includegraphics[width=1\linewidth]{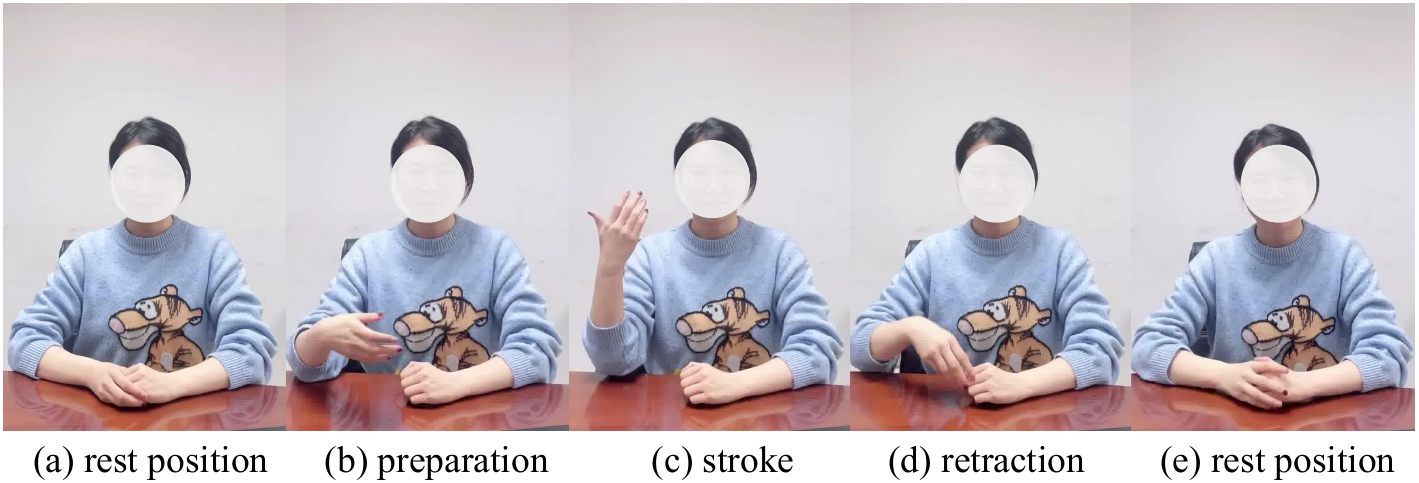}
\end{center}
\caption{Illustration of Gesture Clip in Gesture Lexicon. Gestures typically go through complete stages, including rest pose, preparation, stroke, retraction, and return to rest pose. However, not all Gesture clips have all three stages of preparation, stroke, and retraction.}
\label{fig:2}
\end{figure}

\textbf{Gesture Clips:}
The deep learning-based methods are trained in an end-to-end manner and do not consider the completeness of generated gestures, as mentioned above, particularly in terms of complete stages. However, complete gestures have the potential to enhance the professional and detailed auxiliary expressive effect. To address this, GesGPT model generates gestures with complete stages and professionalism. We primarily extract gesture clips through two approaches: utilizing a publicly available dataset BEAT \cite{36} and collecting the ZHUBO dataset obtained from internet videos. The specific introduction of these datasets will be presented in Section 4.

To extract gesture clips, we first simplify the representation of gestures using the positions of four skeletal points: the left elbow $E_{L}$, right elbow $E_{R}$, left wrist $W_{L}$, and right wrist $W_{R}$ of the human body. This simplified representation is denoted as $D \in \mathbb{R}^{4 \times 3}$ , where in the ZHUBO dataset, the keypoints are represented using three-dimensional spatial positions denoted as $(x, y, z)$ coordinates. In the BEAT and motion capture datasets, the keypoints are represented using euler angles denoted as roll, pitch, and yaw. 
Next, we calculate the distance between adjacent frames $i$ and $i-1$ as $D_{i}-D_{i-1}$, and based on a pre-defined threshold, we filter out the starting and ending points of the gestures, thus obtaining the gesture clip. We found that this detection method performs well in extracting gesture clips in situations where gestures and static rest pose occur intermittently, such as in the ZHUBO dataset. However, it yields unsatisfactory results for continuous motion of a person's gestures, as in the case of the BEAT dataset. Thus, we further optimized the quality of the extracted gesture clips by incorporating partial manual annotation.

\textbf{Gesture Prompt:}
Based on our analysis of text-gesture pairs in the ZHUBO dataset, we have identified some commonalities between the visual appearance of gestures and their intended meanings. For example, explanatory gestures are often accompanied by upward or outward movements of the palm, while emphatic gestures are often accompanied by downward chopping movements of the hand. In order to enhance the functionality of our gesture lexicon, we applied text parsing methods to annotate each gesture unit with its corresponding action intention and semantic word, based on the accompanying text. Subsequently, we enlisted the expertise of a professional speaker to manually refine the original gesture labels in our dataset, thus creating our final semantically enriched gesture library. Each gesture clip is assigned an intention label, and some clips are associated with semantic words, collectively serving as prompts for the gesture clips. 
The example of the gesture lexicon is shown in Fig.4.
\begin{figure}[htbp]
\begin{center}
\includegraphics[width=1\linewidth]{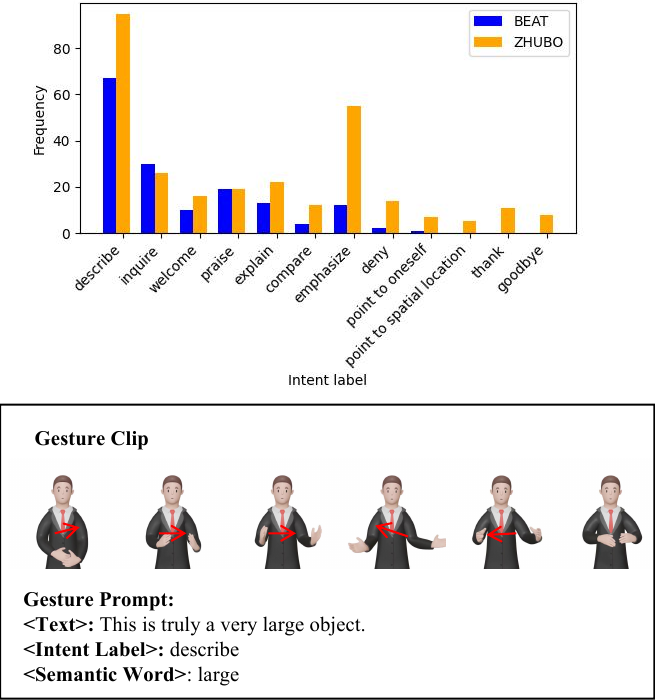}
\end{center}
\caption{Professional Gesture Lexicon. The upper graph represents the quantity of gestures belonging to different intent categories. Below is an example of a gesture clip from the lexicon.} 
\label{fig:3}
\end{figure}

\textbf{Rule-based Searching Method:}
After obtaining the text parsing results, represented as $\{T: <intention:label\_1>,<stroke:label\_2>,<semantic:label\_3>\}$, we employ a linear search algorithm to search for matching keywords in the gesture lexicon based on the $label\_3$ in $<semantic:label\_3>$. If no corresponding gesture prompt is found, we perform a search based on the $label\_1$ in T: $<intention:label\_1>$ and return the corresponding gestures. It is important to note that multiple gestures may be found during the search process. In this case, we randomly select one and integrate it based on the position of $label\_2$ in the text provided by $<stroke:label\_2>$. Additionally, the text and audio are aligned, and the time of the audio can be found based on the position of the text.

\subsubsection{Gesture Integration}
In the field of gesture linguistics, the work by \cite{20} suggests that gestures typically comprise a set of basic gesture units and random perturbations. In this study, we adopt the professional gesture defined in the gesture lexicon as the fundamental gesture unit. Additionally, we incorporate base gestures to introduce random fluctuations and enhance the authenticity of the generated gestures. For example, even in the absence of explicit gestures, the body may exhibit rhythmic oscillations. Subsequently, professional gestures with base gestures are merged to generate the final results.

\textbf{Base Gesture:}
Kucherenko et al. \cite{19} have shown that deep learning methods can effectively capture the association between audio and movement rhythms. In this study, we adopt a learning-based framework to model rhythmic body sway as a base gesture. We employ a one-dimensional convolutional model in the form of a U-NET for base gesture modeling. The base gesture $G^{B}$ is generated under supervision by minimizing the L1 distance between the ground truth $G_{GT}^{B}$ and the predicted gesture $G_{i}^{B}$. To reduce the jitter in the learned gestures, we incorporate a loss function based on higher-order derivatives, as proposed in \cite{37} in (1).We utilize one-Euro algorithm filter \cite{38} to smooth the generated results, obtaining the final base gesture.

$$
  L_{Base}=\frac{1}{N}  \sum_{i=1}^N \left| G_{GT}^{B} -G_{i}^{B}\right|+\\
  \left| G_{GT}^{{B}^{'}} -G_{i}^{{B}^{'}}\right|+
  \left| G_{GT}^{{B}^{"}} -G_{i}^{{B}^{"}}\right|
  \eqno{(1)}
$$

 For the blending between professional gestures $G^{P}$ and base gestures $G^{B}$, we employ linear interpolation. Assuming the total number of frames for the current $G^{P}$ is $N$, we conduct linear interpolation between the $j$ frames preceding $G^{P}$ and the $j$ frames following its end. Here, $j$ represents the interpolation step size, which is set to 6 for our experiment. For the front section fusion of $G^{P}$, the equation is represented as in (2):
 $$
 \begin{aligned}
 G_{i-m}=\frac{1}{m}G^{P}_{1}+(1-\frac{1}{m})G^{B}_{i}
 \end{aligned}
 \eqno{(2)}
  $$
 where $m=(1,2,\ldots,j)$ and $G^{P}_{1}$ represents the first frame of the current $G^{P}$, with $i$ denoting the current frame number. 
 
 For the back section fusion of $G^{P}$, it is expressed as shown in (3):
  $$
 \begin{aligned}
 G_{i+m} = \frac{1}{m}G^{P}_N + (1-\frac{1}{m})G^{B}_{i+m}
  \end{aligned}
 \eqno{(3)}
  $$
 where $G^{P}_N$ represents the last frame of the current $G^{P}$ . We only blend the base gesture with the frames before and after the professional gesture, and do not blend it during the frames of the professional gesture. For motion capture data such as BERT \cite{36}, we convert euler angles to quaternions for linear interpolation.

\section{Experiments}

\subsection{Experimental Data}
We conducted experiments using the BEAT dataset \cite{36} and our proposed ZHUBO dataset, focusing on English and Chinese languages, respectively. For the BEAT dataset, we selected a segment featuring the character `wayne' from the English data and divided it into 35 videos for the training set and 8 videos for the testing set. We obtained the corresponding transcripts for these videos and utilized text parsing and gesture lexicon modules to obtain the parsed text results and annotated professional gestures. Additionally, we used the MFA tool \cite{39} to align the audio and text in the dataset, enabling us to obtain temporal information of the text, which facilitated the fusion of gestures with the speech content.

We collected Chinese host speech videos from the internet, which includes speech videos from multiple persons accompanied by professional gestures. In this paper, we selected segments featuring the character `kanghui' and divided them into 68 videos for the training set and 18 videos for the testing set. We used Mediapipe \cite{40} to obtain the skeletal positions of human bodies in the videos. Following the same methodology mentioned earlier, we obtained text with time annotations from the videos, along with parsed text results and annotated professional gestures. It is worth noting that the annotations for the ZHUBO dataset are in Chinese, and the text input from the ZHUBO dataset used in ChatGPT is also in Chinese.

\subsection{Experimental Evaluation}

\subsubsection{Objective evaluation}
In our study, we performed an evaluation of the GesGPT and benchmarks on two datasets, BEAT and ZHUBO, respectively. For evaluation metrics,  we utilized the Beat Alignment Score introduced in CaMN \cite{36}, which is designed to evaluate the association between gesture motion and speech. Specifically, the score is defined by computing the average distance between the motion beat of each gesture $B_{i}^{G}$ and the closest speech beat $B_{i}^{A}$, as in (4). We adopted the parameter $\sigma=0.3$ setting described in \cite{36} . In the ZHUBO dataset, the speech data is accompanied by background music.
$$
  BeatAlign=\frac{1}{N}\sum_{i=1}^{N}exp(\frac{-min\left | B^{G}_{i}-B^{A}_{i} \right |^{2}}{2\sigma^{2}})
   \eqno{(4)}
$$

Furthermore, we utilized the Fréchet Gesture Distance (FGD) distance proposed in \cite{41}, which stands for Fréchet distance for the Gaussian mean and covariance of latent features, to quantify the similarity between two gestures' distribution features.These distribution features are defined by their mean and variance denoted by $(\mu_{1},\sigma_{1})$ and $(\mu_{2},\sigma_{2})$, respectively,where the gesture distribution features were extracted using a pre-trained model. Specifically, we utilized the FDG distance to evaluate the similarity between generated gestures and real gestures, as in (5). We extracted the distributions using the model pre-trained on the BEAT dataset in \cite{36}, which was only tested on the BEAT dataset and not evaluated on the ZHUBO dataset.

$$
  FGD=\left| \mu_{1} -\mu_{2}\right|^{2}+Tr(\mu_{1}+\mu_{2}-(\mu_{1}\mu_{2})^{\frac{1}{2}})
   \eqno{(5)}
$$

For the BEAT dataset, we utilized the CaMN approach \cite{36} as our baseline. This approach utilizes a sequential model to generate gestures. Regarding the ZHUBO dataset, we employed a baseline network that was trained on a bidirectional gated recurrent unit model \cite{42}. The objective evaluation results are presented in TABLE I. GesGPT outperforms both baselines in terms of rhythmic consistency and realism. This method mitigates the issue of averaging effects that can arise from end-to-end network training.

\begin{table}[h]
\caption{Objective Evaluation Results}
\label{table_example}
\centering
\begin{tabular}{c @{\hspace{0.5cm}} c @{\hspace{0.5cm}} c @{\hspace{0.5cm}} c}
\toprule
Dataset & Method & BeatAlign↑ & FGD↓ \\
\midrule
 & CaMN(baseline)\cite{36} & 0.788 & 183.3 \\
BEAT&DiffuseStyleGesture\cite{21} & 0.789 & 178.19 \\
&QPGesture\cite{22}& 0.790 & 195.14 \\
 & GesGPT & \textbf{0.838} & \textbf{173.09} \\
 \midrule
 & Seq2Seq(baseline)\cite{42} & 0.859 & / \\
ZHUBO& Trimodal\cite{41} & 0.862 & / \\
& Template-BP\cite{3} & 0.870 & / \\
 & GesGPT & \textbf{0.882} & / \\
\bottomrule
\end{tabular}
\begin{flushleft}

\end{flushleft}
\end{table}
\subsubsection{User Study}

We conducted a user study using the BEAT and ZHUBO datasets. From the test sets, we selected 8 videos, each approximately 20 seconds long and containing complete sentences. Users were assigned to evaluate gesture videos of Ground Truth, baseline, and GesGPT in terms of synchrony, naturalness, and semantic expressiveness of gestures, respectively. Ratings for Synchrony, Naturalness, and Semantic were collected on a scale of 1-5, with 1 indicating the lowest quality and 5 indicating the highest quality. We gathered feedback from 50 users for the BEAT dataset and 35 users for the ZHUBO dataset. The subjective evaluation results are presented in Table II. We observed that GesGPT performed superiorly across all three aspects. GesGPT achieved semantic scores close to the ground truth on the BEAT dataset, and its superiority was even more pronounced on the ZHUBO dataset. This could be attributed to the fact that the character in the ZHUBO Chinese dataset predominantly remained in a resting pose state, while in the BEAT English dataset, the characters' gestures exhibited continuous variation. Additionally, the lower quality of video pose estimation in ZHUBO also affects the subjective results. Subjective experiment indicate that our method is capable of generating natural gestures that effectively convey intentions.

\begin{table}[h]
\caption{Subjective Evaluation Results}
\label{table_example}
\centering
\begin{tabular}{c c c c c}
\toprule
Dataset & Method & Synchrony↑ & Naturalness↑ & Semantic↑ \\
\midrule
& GT& \textbf{3.95} & \textbf{3.83} & \textbf{3.79} \\
BEAT & CaMN\cite{36} & 3.54 & 3.50 & 3.42 \\
     & GesGPT & 3.74 & 3.70 & 3.77 \\
\midrule
& GT& 3.29 & 3.36 & 3.21 \\
ZHUBO & Seq2Seq\cite{42} & 3.16 & 3.14 & 3.02 \\
      & GesGPT & \textbf{3.65} & \textbf{3.45} & \textbf{3.40} \\
\bottomrule
\end{tabular}
\begin{flushleft}


\end{flushleft}
\end{table}

\subsection{Visualization Results}

We presented the visualization results of GesGPT and the baseline model on the BEAT dataset, as depicted in Fig.5. The figure illustrates the effectiveness of both methods in generating gestures for complete sentences. To highlight the changes in gestures, we performed frame sampling on the original videos. The green segments represent the positions where the emphasized words appear in the parsed script. As research indicates that gestures often precede co-expressive speech \cite{43}, we synthesized the selected gestures at 3 frames before the emphasized words using GesGPT, while the baseline model generated gestures end-to-end using a sequence model. It is evident that GesGPT is capable of generating more diverse and comprehensive gestures, whereas the baseline model's generated results are relatively monotonous. Language models such as ChatGPT exhibit superior abilities in analyzing sentence expressions, and we propose leveraging these capabilities to generate gestures that offer enhanced auxiliary expressiveness. By creating a comprehensive and specialized gesture lexicon and employing text parsing methods, we can generate high-quality gestures that possess expressive richness.
\begin{figure}[htbp]
\begin{center}
\includegraphics[width=1\linewidth]{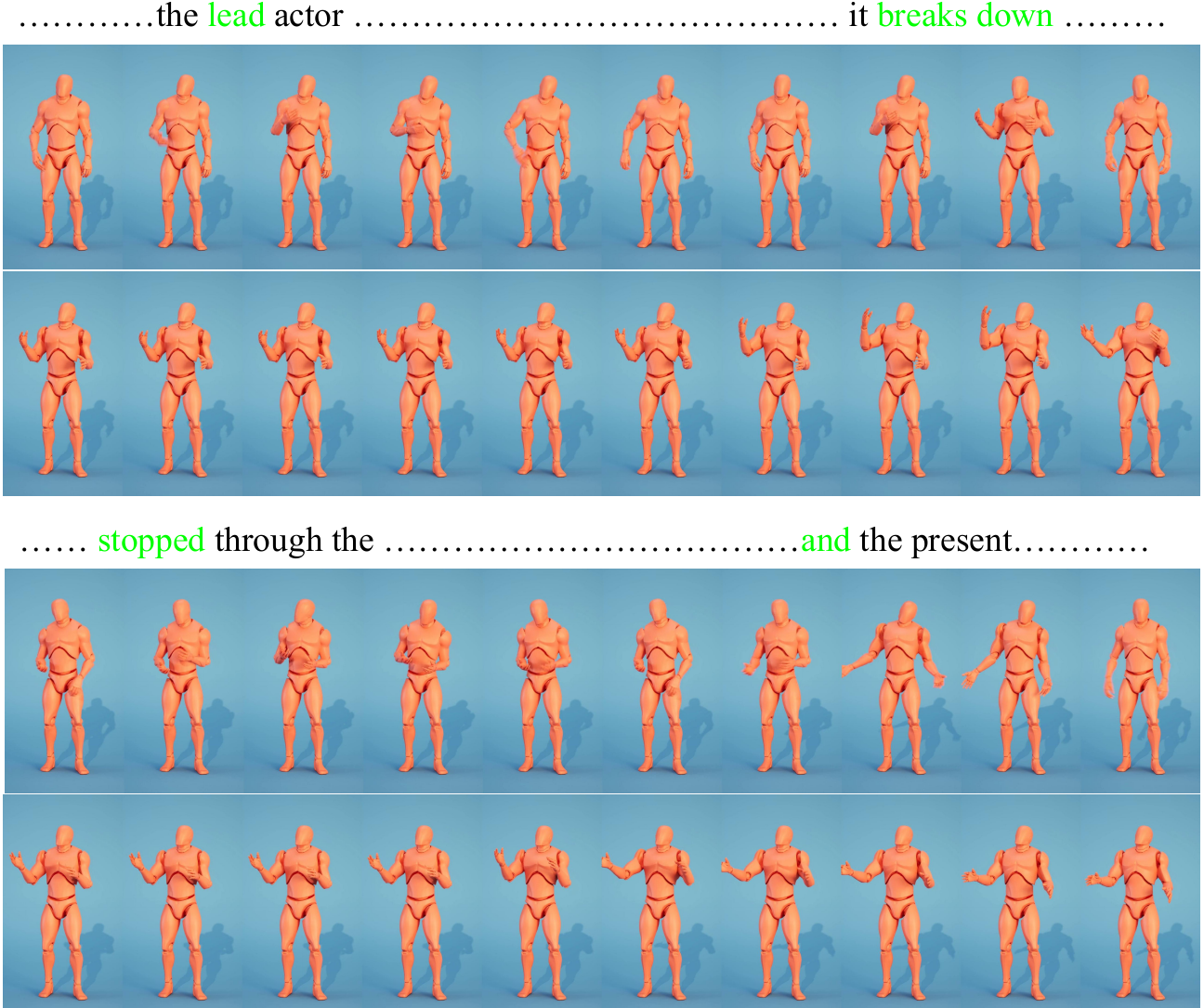}
\end{center}
\caption{Visualization of Generated Examples on the BEAT Dataset. The first line below each text represents the generated results of GesGPT, and the second line represents the generated results of the baseline.}
\label{fig:5}
\end{figure}

We utilized BEAT videos as test data and employed text parsing methods to obtain action scripts. The search strategy and learning-based models described in Section 3 were utilized to generate corresponding gestures. The fused visualization results are shown in Fig.6, where we assume a one-to-one mapping between sentences and intentions, meaning that each sentence can generate a professional gesture. LLMs possess powerful text semantic analysis capabilities, making them suitable for assisting in generating gestures that are more expressive. However, LLMs currently lack direct audio perception. Therefore, we propose utilizing deep learning-based models to learn basic rhythmic gestures from a large amount of data and integrate them with meaningful gestures, providing a more optimized approach. Furthermore, this form of gesture generation based on action scripts enhances controllability, allowing targeted edits to be made to the generated results by modifying the annotations in the script.

\begin{figure}[htbp]
\begin{center}
\includegraphics[width=1\linewidth]{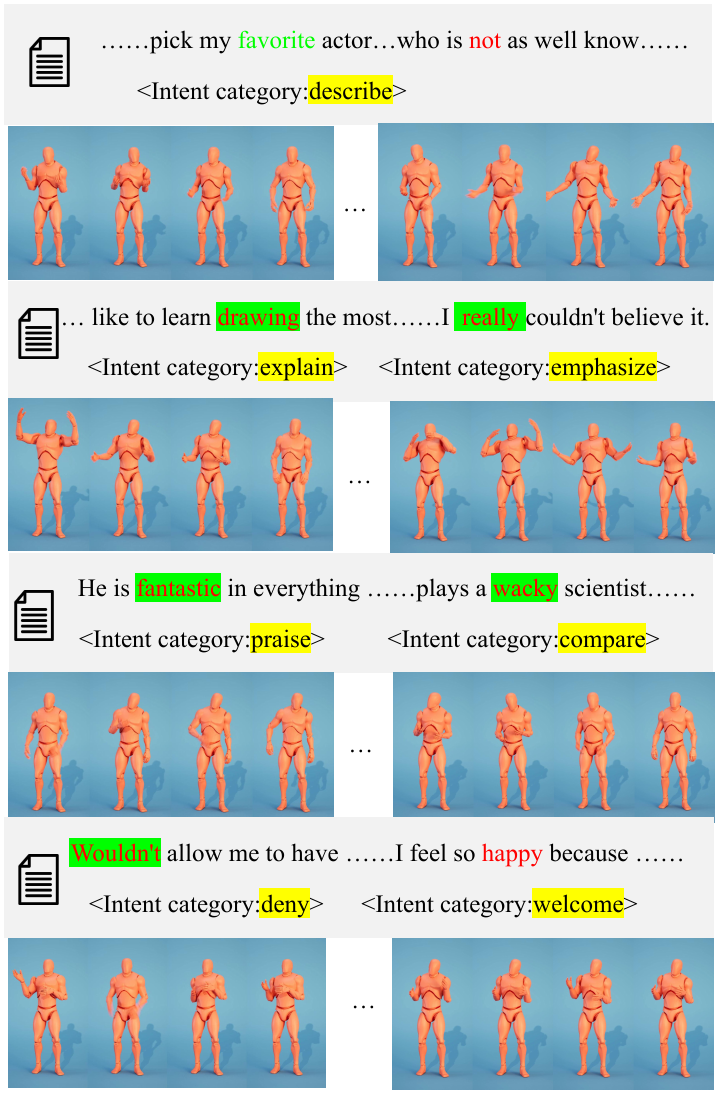}
\end{center}
\caption{Visualization Results of GesGPT. Under the guidance of the text parsing script, GesGPT enables controlled generation of final gestures. Emphasis words are denoted in green, while semantic words are represented in red in the text. Additionally, the intent categories of the statements are highlighted in yellow.}
\label{fig:6}
\end{figure}
\section{Conclusion}
In summary, we propose a new method for semantic co-speech gesture generation based on LLMs, called GesGPT. We use ChatGPT for text parsing to obtain the textual intent, emphasis words, and semantic words, and form a parsing script. Subsequently, we generate and fuse gestures based on the parsing script, leveraging a constructed annotated gesture lexicon. Experimental results on the BEAT and ZHUBO datasets demonstrate that our method can generate natural and expressively rich gestures. By effectively utilizing the capabilities of large-scale language models in text analysis and designing appropriate prompts, our method produces gestures with enhanced intentional meaning and adaptability to context. This research highlights the potential of ChatGPT in embodied intelligence and gesture synthesis, showcasing its effectiveness in generating meaningful gestures. We believe that further improvements in gesture generation can be achieved through enhanced semantic analysis of text input. 

However, our study has several limitations. Firstly, we made the assumption that each sentence contains only one intent during text parsing, which may restrict the capability to express multiple intents in longer texts or texts without specific intents. Additionally, considering more contextual information beyond a single sentence during text parsing could potentially improve parsing effectiveness. Moreover, the gesture lexicon constructed in this study was based on data from specific speakers, reflecting individual styles. In future research, we aim to expand upon this work by enriching the parsing results and creating a more generalized gesture dataset.






\section*{ACKNOWLEDGMENT}

This work was supported by the National Key R\&D Program of China (2022YFF0902202).


\end{document}